\documentclass[12pt,pdftex,noinfoline]{imsart}

\usepackage[usenames,dvipsnames]{xcolor}
\usepackage{natbib}
\bibpunct{(}{)}{;}{a}{,}{,}

\usepackage[pdftex,pdfborderstyle={/S/U/W 1},colorlinks]{hyperref}
\RequirePackage[OT1]{fontenc}
\usepackage{algorithm}
\usepackage{algorithmic}
\RequirePackage{amsthm,amsmath,amsfonts}
\RequirePackage{hypernat}
\usepackage{fullpage}
\usepackage{graphicx}
\graphicspath{ {fig/} }
\usepackage{hyperref}
\usepackage{amsthm, amssymb, amsmath}
\usepackage{url}
\usepackage[english]{babel}
\usepackage{times}
\usepackage[T1]{fontenc}
\usepackage{bbm}
\usepackage{xspace}
\usepackage{rotating,multirow}
\usepackage{enumerate}
\usepackage{mathtools}
\usepackage{booktabs}
\usepackage{caption}
\usepackage{subcaption}

\usepackage{tikz}
\usetikzlibrary{shapes}
\usepackage{tkz-euclide}
\usepackage{ifthen}
\usetikzlibrary{arrows,positioning}

\DeclareMathOperator*{\argmin}{\arg\min}

\def\interleave{|\kern-.25ex|\kern-.25ex|}
\def\interleavesub{|\kern-.15ex|\kern-.15ex|}

\newcommand{\nNorm}[1]{\left|\kern-.25ex\left|\kern-.25ex\left| {#1}\right|\kern-.25ex\right|\kern-.25ex\right|}

\newcommand{\R}{{\mathcal R}}

\def\min{\mathop{\text{\rm min}}}
\def\max{\mathop{\text{\rm max}}}

\numberwithin{equation}{section}
\theoremstyle{plain}

\newtheorem{lemma}{Lemma}[section]
\newtheoremstyle{remark}{\topsep}{\topsep}%
     {\normalfont}
     {}           
     {\bfseries}  
     {.}          
     {.5em}       
     {\thmname{#1}\thmnumber{ #2}\thmnote{ #3}}
\theoremstyle{remark}

\long\def\comment#1{}
\def\reals{{\mathbb R}}

\def\supp{\mathop{\text{supp}\kern.2ex}}
\def\argmin{\mathop{\text{\rm arg\,min}}}

\let\hat\widehat
\let\tilde\widetilde

\let\hat\widehat
\let\tilde\widetilde

\def\1{{(1)}}
\def\2{{(2)}}

\long\def\comment#1{}

\long\def\comment#1{}
\def\reals{{\mathbb R}}

\def\supp{\mathop{\text{supp}\kern.2ex}}
\def\argmin{\mathop{\text{\rm arg\,min}}}

\let\tilde\widetilde

\let\hat\widehat
\let\tilde\widetilde

\def\1{{(1)}}
\def\2{{(2)}}

\long\def\comment#1{}

\def\threebars{\mbox{$|\kern-.25ex|\kern-.25ex|$}}

\begin{document}

\hypersetup{citecolor=MidnightBlue}
\hypersetup{linkcolor=Black}
\hypersetup{urlcolor=MidnightBlue}

\setlength{\parskip}{0.5em}

\begin{frontmatter}
\begin{center}
\Large\bf Prediction Rule Reshaping
\end{center}
\runtitle{Prediction Rule Reshaping}
\begin{aug}
\vskip3pt
\author{\fnms{Matt} \snm{Bonakdarpour\thanksref{chi}}\ead[label=e1]{mbonakda@gmail.com}}
\; \author{\fnms{Sabyasachi} \snm{Chatterjee\thanksref{ill}}\ead[label=e1]{sc1706@illinois.edu}}
\; \author{\fnms{Rina} \snm{Foygel Barber\thanksref{chi}}\ead[label=e1]{rina@galton.uchicago.edu}}
\; \author{\fnms{John} \snm{Lafferty\thanksref{yale}}\ead[label=e2]{john.lafferty@yale.edu}}
\thankstext{chi}{Department of Statistics, The University of Chicago}
\thankstext{ill}{Department of Statistics, University of Illinois at Urbana-Champaign}
\thankstext{yale}{Department of Statistics and Data Science, Yale University}
\affiliation{Some University and Another University}
\address{
{\normalsize \today}
}
\end{aug}
\begin{abstract}
Two methods are proposed for high-dimensional shape-constrained regression and classification.  
These methods \textit{reshape} pre-trained prediction rules to satisfy shape constraints like 
monotonicity and convexity. The first method can be applied to any pre-trained prediction rule, while 
the second method deals specifically with random forests. In both cases, efficient algorithms are 
developed for computing the estimators, and experiments are performed to demonstrate their 
performance on four datasets. We find that reshaping methods enforce shape constraints 
without compromising predictive accuracy.
\end{abstract}

\vskip20pt 
\end{frontmatter}

\section{Introduction} \label{sec:intro}
Shape constraints like monotonicity and convexity arise naturally in
many real-world regression and classification tasks. For example,
holding all other variables fixed, a practitioner might assume that
the price of a house is a decreasing function of neighborhood crime rate, 
that an individual's utility function is concave in income level, or that
phenotypes such as height or the likelihood of contracting a disease
are monotonic in certain genetic effects.

Parametric models like linear regression implicity impose monotonicity constraints at the cost of 
strong assumptions on the true underlying function. On the other hand, nonparametric techniques like 
kernel regression impose weak assumptions, but do not guarantee monotonicity or 
convexity in their predictions. Shape-constrained nonparametric regression methods attempt to offer 
the best of both worlds, allowing practitioners to dispense with parametric assumptions while retaining 
many of their appealing properties. 

However, classical approaches to nonparametric regression under shape constraints suffer from the 
curse of dimensionality \citep{han2016,han2017}. Some methods have been developed to mitigate 
this 
issue under assumptions like additivity, where the true 
function $f$ is assumed to have the form $f(x) = \sum_j f_j(x_j) + c$, where a subset of the component
$f_j$'s are shape-constrained \citep{pya2015,chen2016,xu2016}. But in many real-world settings, the 
lack 
of interaction terms among the predictors can be too restrictive.

Approaches from the machine learning community like random forests, gradient boosted trees, and 
deep learning methods have been shown to exhibit outstanding empirical performance on 
high-dimensional tasks. But these methods do not guarantee monotonicity or convexity. 

In this paper, we propose two methods for high-dimensional shape-constrained regression and 
classification. These methods blend the performance of machine learning 
methods with the classical least-squares approach to nonparametric 
shape-constrained regression.

In Section~(\ref{sec:bb}), we describe black box reshaping, which takes any pre-trained prediction 
rule and reshapes it on a set of test inputs to enforce  shape constraints. In the case of monotonicity
constraints, we develop an efficient algorithm to compute the estimator. Section~(\ref{sec:rrf}) 
presents a second method designed specifically to reshape random forests 
\citep{breiman2001}. 
This approach reshapes each 
individual decision tree based on its split rules and estimated leaf values. Again, in the 
case of monotonicity constraints, we present another efficient reshaping algorithm. We apply our 
methods to four datasets in Section~(\ref{sec:exp}) and show that they enforce the 
pre-specified shape constraints without sacrificing accuracy. 

\subsection{Related Work}
In the context of monotonicity constraints, the black box reshaping method is related to the method of 
rearrangements \citep{chernozhukov2009,chernozhukov2010}. The rearrangement operation 
takes 
a pre-trained prediction rule and  sorts its predictions to enforce monotonicity. In higher 
dimensions, the rearranged estimator is the average of one-dimensional rearrangements. In contrast, 
this paper focuses on isotonization of prediction values, jointly reshaping multiple dimensions in 
tandem. It would be interesting to explore adaptive procedures that average rearranged and isotonized 
predictions in future work.

Monotonic decision trees have previously been studied in the context of classification. Several 
methods require that the training data satisfy monotonicity constraints 
\citep{potharst2002,makino1996}, a relatively strong assumption in the presence of noise. The 
methods 
we 
propose here do not place any restrictions on the training data. 

Another class of methods augment the score function for each split to incorporate the degree of 
non-monotonicity introduced by that split \citep{ben-david1995,gonzalez2015}. However, this 
approach does not guarantee monotonicity. \citet{feelders2003} apply pruning algorithms to 
non-monotonic trees as a post-processing step in order to enforce monotonicity. For a 
comprehensive survey of estimating monotonic functions, see \citet{gupta2016} .

A line of recent work has led to a method for 
learning deep monotonic models by alternating different types of monotone layers \citep{you2017}. 
\citet{amos2017} propose a method for fitting neural networks whose predictions are convex 
with respect to a subset of predictors. 

Our methods differ from this work in several ways. First, our techniques can be used to enforce  
both monotonic and convex/concave relationships. Unlike pruning methods, neither approach 
presented here changes the structure of the original tree. Black box reshaping, described in 
Section~(\ref{sec:bb}), can be applied to any pre-trained prediction rule, giving practitioners the 
flexibility of picking the method of their choice. And both methods guarantee that the 
intended shape constraints are satisfied on test data. 

\section{Prediction Rule Reshaping}
In what follows, we say that a function $f: \reals^d \rightarrow \reals$  is monotone with respect to 
variables $\mathcal{R} \subseteq [d] = \{1,\dots,d\}$ if $f(x) \leq f(y)$ when $x_i \leq y_i$ for $i \in 
\mathcal{R}$, and 
$x_i = y_i$ otherwise. 

Similarly, a function $f$ is convex in
$\mathcal{R}$ if for all $x,y \in \reals^d$ and $\alpha \in [0,1]$, $f(\alpha x + (1-\alpha)y) \leq \alpha f(x) 
+ (1-\alpha)f(y)$ 
when 
$x_i = y_i 
\; \forall i \notin \mathcal{R}$.

\subsection{Black Box Reshaping}
\label{sec:bb}
Let $\hat{f}~:~\mathbb{R}^d \rightarrow \mathbb{R}$ denote an arbitrary
prediction rule fit on a training set and assume we have a candidate set of 
shape constraints with respect to variables $\R \subseteq [d]$. For example, we might require 
that the function be monotone increasing in each variable $v \in \mathcal{R}$. 

Let $\mathcal{F}$ denote the class of functions that satisfy the 
desired shape 
constraints on each 
predictor 
variable $v \in \mathcal{R}$. We aim to find a function $f^\ast \in \mathcal{F}$ that is close to $\hat{f}$ 
in the $L_2$ norm:
\begin{equation} \label{eq:bb-inf}
f^\ast = \argmin_{f \in \mathcal{F}} \lVert f - \hat{f} \rVert_2
\end{equation}
where the $L_2$ norm is with respect to the uniform measure on a compact set containing the data. 
We simplify this infinite-dimensional problem by only considering 
values of $\hat{f}$ on certain fixed test points. 

Suppose we take a sequence $t^1, t^2, \dots, t^n$ of test points, each in $\reals^d$, that 
differ only in their $v$-th coordinate so that $t^i_k=t^{i'}_k$ for all $k\neq v$. These points can be 
ordered by their $v$-th coordinate, allowing us to consider shape constraints on the vector $(f(t^1), 
f(t^2), ..., f(t^n)) \in \reals^n$. For instance,  under a
monotone-increasing constraint with respect to $v$, if $t^1_v \leq t^2_v \leq \dots \leq t^n_v$, then we 
consider functions $f$ such that $(f(t^1), 
f(t^2), ..., f(t^n))$  is a monotone sequence.

There is now the question of choosing a test point $t$ as well as a sequence of 
values $t^1_v, ... , t^n_v$ to plug into its $v$-th coordinate. A natural choice is to 
use the observed data values as both test points and coordinate values.

Denote $\mathcal{D}_n = \{(x^1,y^1),\dots,(x^n,y^n)\}$ as a set of observed values where $y^i$ is the 
response and $x^i \in \reals^d$ are the predictors. From each $x^i$, we 
construct 
a sequence of test points that can be ordered according to their $v$-th coordinate in the 
following way. Let $x^{i,k,v}$ denote the observed vector $x^i$ with its $v$-th coordinate replaced by 
the $v$-th coordinate of $x^k$, so that
\begin{equation} 
x^{i,k,v} = (x^i_1, x^i_2, \ldots, x^i_{v-1},x^k_v, x^i_{v+1}, \ldots, x^i_d).
\end{equation}
This process yields $n$ points from $x^i$ that can be ordered by their $v$-th coordinate, 
$x^{i,1,v},x^{i,2,v},\dots,x^{i,n,v}$. We then require  
$(f(x^{i,1,v}),f(x^{i,2,v}),\dots,f(x^{i,n,v}))\in S_v$ where  $S_v \subset \reals^d$ 
is the appropriate convex cone that enforces the shape constraint for variable $v \in \mathcal{R}$, for 
example the cone of monotone 
increasing or convex sequences.

\pagebreak

To summarize, for each coordinate $v \in \mathcal{R}$ and for each $i \in [n]$, we:
\begin{enumerate}
	\item Take the $i$-th observed data point $x^i$ as a test point.
	\item Replace its $v$-th coordinate with the $n$ observed $v$-th coordinates $x^1_v, ... x^n_v$ to 
	produce\\ $x^{i,1,v},x^{i,2,v},\dots,x^{i,n,v}$.
	\item Enforce the appropriate shape constraint on the vector of evaluated function values,\\ 
	$(f(x^{i,1,v}),f(x^{i,2,v}),\dots,f(x^{i,n,v}))\in S_v$. 
\end{enumerate}
See Figure~(\ref{fig:bb}) for an illustration. This leads to the following relaxation of (\ref{eq:bb-inf}): 
\begin{figure}[t!]
	\centering
	\begin{tikzpicture}[real/.style={circle,draw,fill=white!40,minimum 
		size=20},synthetic/.style={circle,draw,fill=gray!20,very thick,minimum size=20}]
	\tikzstyle{every node}=[font=\small]
	\foreach \x[count=\xi from 1] in {0,...,3}
	\foreach \y[count=\yi from 1]  in {0,...,3} 
	{\pgfmathtruncatemacro{\label}{\x - 5 *  \y +21}
		\def\lbl{$x$}
		\tkzInit[xmax=5,ymax=5,xmin=0,ymin=0]
		\tkzDrawX
		\tkzDrawY
		
		\ifthenelse{\x=0 \AND \y=0}{\def\mycol{synthetic}}{\def\mycol{real}}	
		\ifthenelse{\x=0 \AND \y=0}{\def\lbl{$x^{1,1}$}}{}
		\ifthenelse{\x=0 \AND \y=1}{\def\lbl{$x^{1,3}$}}{}
		\ifthenelse{\x=0 \AND \y=2}{\def\lbl{$x^{1,4}$}}{}
		\ifthenelse{\x=0 \AND \y=3}{\def\lbl{$x^{1,2}$}}{}
		
		\ifthenelse{\x=1 \AND \y=3}{\def\mycol{synthetic}}{}
		\ifthenelse{\x=1 \AND \y=0}{\def\lbl{$x^{2,1}$}}{}
		\ifthenelse{\x=1 \AND \y=1}{\def\lbl{$x^{2,3}$}}{}
		\ifthenelse{\x=1 \AND \y=2}{\def\lbl{$x^{2,4}$}}{}
		\ifthenelse{\x=1 \AND \y=3}{\def\lbl{$x^{2,2}$}}{}
		
		\ifthenelse{\x=2 \AND \y=1}{\def\mycol{synthetic}}{}
		\ifthenelse{\x=2 \AND \y=0}{\def\lbl{$x^{3,1}$}}{}
		\ifthenelse{\x=2 \AND \y=1}{\def\lbl{$x^{3,3}$}}{}
		\ifthenelse{\x=2 \AND \y=2}{\def\lbl{$x^{3,4}$}}{}
		\ifthenelse{\x=2 \AND \y=3}{\def\lbl{$x^{3,2}$}}{}
		
		\ifthenelse{\x=3 \AND \y=2}{\def\mycol{synthetic}}{}
		\ifthenelse{\x=3 \AND \y=0}{\def\lbl{$x^{4,1}$}}{}
		\ifthenelse{\x=3 \AND \y=1}{\def\lbl{$x^{4,3}$}}{}
		\ifthenelse{\x=3 \AND \y=2}{\def\lbl{$x^{4,4}$}}{}
		\ifthenelse{\x=3 \AND \y=3}{\def\lbl{$x^{4,2}$}}{}

		\node [\mycol]  (\x\y) at (1.25*\x+0.75,1.25*\y+0.75) {\lbl};
	} 
	\tikzset{edge/.style = {->,> = latex'}}
	\foreach \x in {0,...,3}
	\foreach \y [count=\yi] in {0,...,2}  {
		\draw[edge] (\x\y)-- (\x\yi);
	}
	\end{tikzpicture}
	\caption{Two-dimensional illustration of the black box setup when reshaping the $y$ dimension. 
		$x^{i,k}$ denotes the original test point $x^i$ with its $y$-coordinate replaced by the 
		$y$-coordinate of $x^k$. Dark gray nodes represent the original observed points.  For 
		monotonicity constraints, a directed edge from node $x$ to node $y$ represents the constraint 
		$f(x) \leq f(y)$ on the shape-constrained function $f$.} 
	\label{fig:bb}
\end{figure}
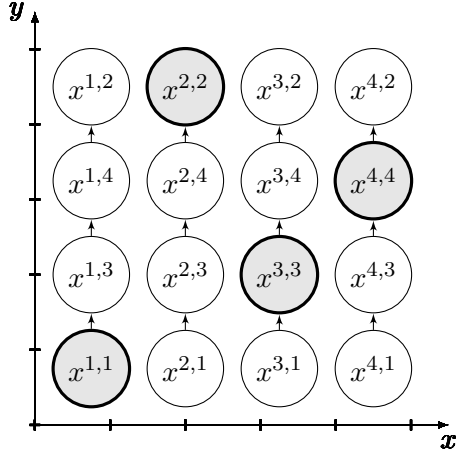

\begin{equation} \label{eq:bb-inf2}
f^\ast = \argmin_{f \in \mathcal{F}_n} \lVert f - \hat{f} \rVert_2 
\end{equation}
where $\mathcal{F}_n$ is the class of functions $f$ such that 
$(f(x^{i,1,v}),f(x^{i,2,v}),\dots,f(x^{i,n,v}))\in S_v \subset \reals^n$ for each $v\in \mathcal{R}$ and each  
$i \in [n]$. In other words, we have relaxed the shape constraints on the function $f$, requiring the
constraints to hold relative to the selected test points. However, this optimization is still 
infinite dimensional. 

We make the final transition to finite dimensions by changing the objective function to only consider 
values of $f$ on the test points. Letting $F_{i,k,v}$ denote the value of $f$ evaluated on test point 
$x^{i,k,v}$, we relax 
(\ref{eq:bb-inf2}) to obtain the solution $F^\ast  = (F^{\ast}_{i,k,v})_{v\in \R,i\in [n],k\in[n] }$ of the 
optimization:
\begin{equation}
\begin{aligned}
&\argmin_F && \sum_{i,k,v} (F_{i,k,v} - 
\hat{f}(x^{i,k,v}))^2  \\
&\text{subject to}&&(F_{i,1,v}, ..., F_{i,n,v}) \in (S_v)_{v \in \mathcal{R}}, \, \forall 
\, 
i 
\in [n]
\end{aligned}
\end{equation}
However, this leads to ill-defined predictions on the original data points $x^1, ..., x^n$, since  for each 
$v$, 
$x^{i,i,v} = 
x^i$, but we may 
obtain different values $F^{\ast}_{i,i,v}$ for various $v\in \R$. 

We avoid this issue by adding a consistency constraint (\ref{eq:consistency}) to obtain our final black 
box reshaping 
optimization (\texttt{BBOPT}):
\begin{alignat}{2}
&\argmin_F&&\quad \sum_{i,k,v} (F_{i,k,v} - \hat{f}(x^{i,k,v}))^2 \\
& \text{subject to} && \quad (F_{i,1,v}, ..., F_{i,n,v}) \in (S_v)_{v \in \mathcal{R}}, \; \forall \, i \in [n] \\
& \text{and} &&\quad F_{i,i,v} = F_{i,i,w} \quad \forall \;v,w\in \mathcal{R}, \forall \, i \in [n] 
\label{eq:consistency}
\end{alignat}
We then take the reshaped predictions to be
\begin{equation*}
f^\ast(x^i) = F^{\ast}_{i,i,v} 
\end{equation*}
for any $v\in \R$.  Since the constraints depend on each $x^i$ independently, \texttt{BBOPT} 
decomposes into $n$ optimization problems, one for each observed value. Note that the true response 
values $y^i$ are not used when 
reshaping. We could select optimal shape constraints on a held-out test set.

\subsubsection{Intersecting Isotonic Regression}
In this section, we present an efficient algorithm for solving
\texttt{BBOPT} for the case when each $\mathcal{S}_v$ imposes monotonicity constraints. Let $R = 
|\mathcal{R}|$ denote the number of monotonicity constraints. 

When reshaping with respect to only one predictor ($R = 1$), the consistency constraints 
(\ref{eq:consistency}) vanish, so the optimization
decomposes into $n$ isotonic regression problems. Each problem 
is efficiently solved in $\Theta(n)$ time with the pool adjacent violators algorithm (PAVA) 
\citep{ayer1955}.

For $R > 1$ monotonicity constraints, \texttt{BBOPT} gives rise to $n$ independent \textit{intersecting 
	isotonic regression} problems. The $k$-th problem corresponds to the $k$-th observed value 
	$x^k$; 
the 
``intersection" is implied by the consistency constraints (\ref{eq:consistency}). For each independent 
problem, our algorithm takes $O(m\log 
R)$ time, where $m = n\times R$ is the number of variables in each problem.

We first state the general problem. Assume $v^1, v^2, \dots, v^K$ are each real-valued vectors with 
dimensions 
$d_1,d_2,\dots,d_K$, respectively.  Let $i_j \in \{1,\dots,d_j\}$ denote an index in the $j$-th vector 
$v^j$. 
The intersecting isotonic regression problem (\texttt{IISO}) is:
\begin{equation}
\begin{aligned}
& \underset{{(\hat{v}^k)_{k=1}^K}}{\text{minimize}} & & \sum_{k=1}^K \lVert \hat{v}{}^k - v^k \rVert^2 \\  
\label{eq:iso}
& \text{subject to} & & \hat{v}^k_1 \leq \hat{v}^k_2 \leq \dots \leq \hat{v}^k_{d_k}, \; \forall \, k \in [K] \\
& \text{and} & & \hat{v}^1_{i_1} = \hat{v}^2_{i_2} = \dots = \hat{v}^K_{i_K}
\end{aligned}
\end{equation}

First consider the simpler constrained isotonic regression problem with a single 
sequence $v~\in~\reals^d$, index $i \in [d]$, and 
fixed 
value $c \in \reals$
\begin{equation}
\begin{aligned}
& \underset{{\hat{v}}}{\text{minimize}} & & \lVert \hat{v} - v \rVert^2 \\  
\label{eq:simple-iso}
& \text{subject to} & & \hat{v}_1 \leq \hat{v}_2 \leq \dots \leq \hat{v}_{d} \\
& \text{and} & & \hat{v}_{i} = c 
\end{aligned}
\end{equation}

\begin{lemma}\label{lem:c-iso}
	The solution $v^\ast$ to (\ref{eq:simple-iso}) can be computed by using index $i$ as a 
	pivot and 
	splitting $v$ into its left and right tails,  so that $\ell = (v_1,v_2,\dots,v_{i-1})$ and $r = (v_{i+1}, 
	\dots, 
	v_d)$, then applying PAVA to obtain monotone tails $\hat{\ell}$ and $\hat{r}$. $v^\ast$ is obtained 
	by 
	setting elements of $\hat{\ell}$ and $\hat{r}$ to
	\begin{equation} \label{eq:min-max}
	\begin{aligned} 
	\ell_k^\ast &= \min (\hat{\ell}_k, c) \\
	r_k^\ast &= \max (\hat{r}_k, c)
	\end{aligned}
	\end{equation}
	and concatenating the resulting tails so that $v^\ast~=~(\ell^\ast, c, r^\ast) \in \reals^d$.
\end{lemma}

\begin{algorithm}[tb]
	\caption{\texttt{IISO Algorithm}}
	\label{alg:iiso}
	\begin{algorithmic}
		\STATE 1. Apply PAVA to each of the 2K tails. 
		\STATE 2. Combine and sort the left and right tails separately.
		\STATE 3. Find segment $s^\ast$ in between tail values where the derivative $g'(\eta)$ 
		changes sign.
		\STATE 4. Compute $c^\ast$, the  minimizer of $g(c)$ in segment $s^\ast$. 
	\end{algorithmic}
\end{algorithm}

We now explain the \texttt{IISO Algorithm} presented in Algorithm~(\ref{alg:iiso}). First 
divide each vector $v^j$ into two tails, the left tail $\ell^j$ and the right tail $r^j$, using the 
intersection index $i_j$ as a pivot,
\begin{align*}
v^j = (\underbrace{v_1^j, v_2^j,\dots, v_{(i_j-1)}^j}_{\ell^j}, v_{i_j}^j, \underbrace{v_{(i_j+1)}^j, 
	v_{(i_j+2)}^j,\dots, v_{d_j}^j}_{r^j}).
\end{align*}
resulting in $2K$ tails $\{\ell^1,\dots,\ell^K,r^1,\dots,r^K\}$. 

Step 1 of Algorithm~(\ref{alg:iiso}) performs an 
unconstrained isotonic 
regression on each tail using PAVA to obtain $2K$ monotone tails 
$\{\hat{\ell}^1,\dots,\hat{\ell}^K,\hat{r}^1,\dots,\hat{r}^K\}$. This can be done in $\Theta(n)$ time, where 
$n$ is the total number of elements across all vectors so that $n = \sum_{i=1}^K d_i$.

Given the monotone tails, we can write a closed-form expression for the 
\texttt{IISO} objective function in terms of the value at the point of intersection.

Let $c$ be the value of the vectors at the point of intersection so that 
$c=\hat{v}_{i_1}^1=\hat{v}_{i_2}^2=\dots=\hat{v}_{i_K}^K$. For a fixed $c$, we can solve 
\texttt{IISO} by applying Lemma~(\ref{lem:c-iso}) to each sequence separately. This yields the following 
expression for the squared error as a function of $c$:
\begin{align}
\begin{split}
g(c) = \sum_{k=1}^K(c &- v_{i_k}^k)^2 + \sum_{k=1}^K \sum_{i=1}^{i_k-1} (\ell_i^k - 
\text{min}(\hat{\ell}_i^k, 
c))^2 + \sum_{l=1}^K \sum_{j=i_l+1}^{d_l} (r_j^l - \text{max}(\hat{r}_j^l, c))^2 
\end{split}
\end{align}
which is piecewise quadratic with knots at each $\hat{\ell}_i^k$ and $\hat{r}_j^l$. Our goal is to find 
$c^\star = \min_c g(c)$. Note that $g(c)$ is convex 
and differentiable.

We proceed by computing the derivative of $g$ at each knot, from smallest to largest, and finding the 
segment in which the sign of the derivative changes from negative to positive. The minimizer $c^\ast$ 
will live in this segment. 

Step 2 of Algorithm~(\ref{alg:iiso}) merges the left and right sorted tails into two sorted lists. This can 
be done in 
$O(n\log K)$ time with a heap data structure. Step 3 computes the derivative of the objective function 
$g$ at each knot, from 
smallest to largest, searching for the segment in which the derivative changes sign. Step 4 
computes the minimizer of $g$ in the corresponding segment. By updating the derivative incrementally 
and 
storing relevant side information, Steps 3 and 4 can be done in linear time. 

The total time complexity is therefore $O(n\log (K))$.
\tikzset{circ/.style={circle, draw, fill=white, scale=1, minimum 
		size=20},sc/.style={circle,draw,fill=gray!40}}
\begin{figure}
	\centering
	\begin{tikzpicture}[thick,scale=0.6, every node/.style={scale=0.8}]
	\node (l1) [sc]{$p_2$};
	
	\node (l2) [circ, below left=2cm of l1]{};
	\node (l3)[circ, below right=2cm of l1]{};
	\draw[-] (l1) to node [above left] {$x_v \leq t_2$} (l2);
	\draw[-] (l1) to node [auto] {$x_v > t_2$} (l3);
	
	\node (l5)[sc, below left=1cm of l2]{$p_1$};
	\node (l4)[below right=1cm of l2]{};
	\draw[-] (l2) to node [auto] {} (l4);
	\draw[-] (l2) to node [auto] {} (l5);
	
	\node (l6)[below left=1cm of l3]{};
	\node (l7)[below right=1cm of l3]{$\ell_3$};
	\draw[-] (l3) to node [auto] {} (l6);
	\draw[-] (l3) to node [auto] {} (l7);
	
	\node (l8)[below left=1cm of l5, scale=.9]{$\ell_1$};
	\node (l9)[below right=1cm of l5, scale=.9]{$\ell_2$};
	\draw[-] (l5) to node [above left] {$x_v \leq t_1$} (l8);
	\draw[-] (l5) to node [auto] {$x_v > t_1$} (l9);
	\end{tikzpicture}
	\caption{An illustration motivating the random forest reshaping method. The only nodes that split on 
		the shape-constrained variable $v$ are $p_1$ and $p_2$. Assume $x\in 
		\reals^d$ drops down to leaf $\ell_1$ and that, holding all other variables constant, increasing 
		$x_v$ beyond $t_1$ and $t_2$, results in dropping to leaves $\ell_2$ and $\ell_3$, respectively. To 
		enforce monotonicity in $v$ for this point, we need to ensure leaf values $\mu_\ell$ follow 
		$\mu_{\ell_1} 
		\leq 
		\mu_{\ell_2} \leq \mu_{\ell_3}$. } 
	\label{fig:rrf}
\end{figure}
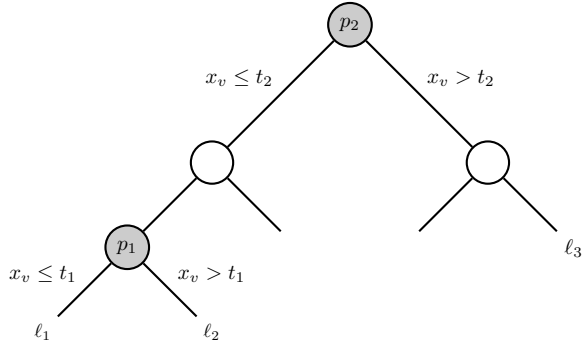

\subsection{Reshaping Random Forests}
\label{sec:rrf}
In this section, we describe a framework for reshaping a random forest to ensure monotonicity of its
predictions in a subset of its predictor variables. A similar method can be applied to ensure convexity. 
For both regression and probability trees \citep{malley2012}, the prediction of the forest is an average 
of 
the prediction of 
each tree; it is therefore sufficient to ensure monotonicity or convexity of the trees.  For the rest of 
this section, we focus on reshaping individual trees to enforce monotonicity. 

Our method is a two-step procedure. The first step is to grow a tree in the usual way. The second step 
is to \textit{reshape} the leaf values to enforce monotonicity. We hope to explore the implications of 
combining these steps in future work. 

Let $T(x)$ be a regression tree and  $\R \subseteq [d]$  a set of
predictor variables to be reshaped. Let $x \in \reals^d$ be an input point and denote the $k$-th 
coordinate of $x$ as $x_k$. Assume $v \in \mathcal{R}$ is a predictor variable to be reshaped. The 
following thought experiment, illustrated in Figure~(\ref{fig:rrf}), will motivate our approach. 

Imagine dropping $x$ down $T$ until it falls in its corresponding leaf, $\ell_1$.  Let $p_1$ be the 
closest ancestor node 
to $\ell_1$ that splits on $v$ and assume it has split rule $\{x_v \leq t_1 \}$. Holding all 
other coordinates constant, increasing $x_v$ until it is greater than $t_1$ would create a new point 
that drops down to a different leaf $\ell_2$ in the right subtree of $p_1$. 

If $\ell_1$ and $\ell_2$ both share another ancestor $p_2$ farther up the tree with split rule $\{x_v \leq 
t_2\}$, increasing $x_v$ beyond $t_2$ would yield another leaf $\ell_3$. Assume these 
leaves have no other shared ancestors that split on $v$. Denoting the  value of leaf $\ell$ as 
$\mu_\ell$, 
in order to ensure monotonicity in $v$ for this point $x$, we require $\mu_{\ell_1} \leq \mu_{\ell_2} \leq 
\mu_{\ell_3}$.

We use this line of thinking to propose a framework for estimating monotonic random 
forests and describe two estimators that fall under this framework. 
\tikzset{circ/.style={circle, draw, fill=white, scale=1, minimum 
		size=20},sc/.style={circle,draw,fill=gray!40}}
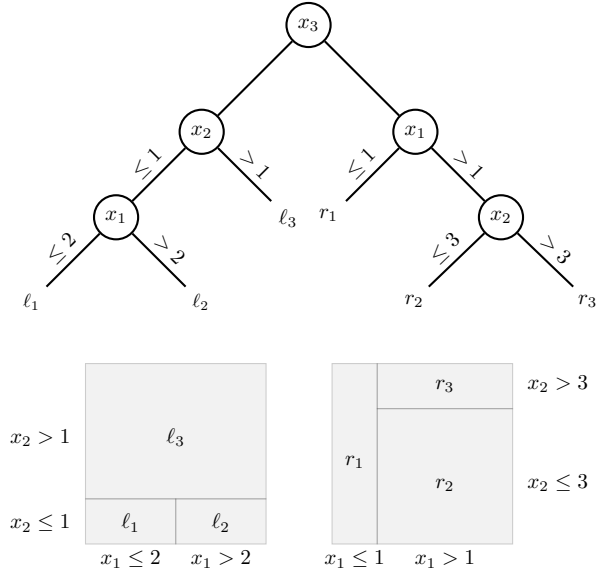
\begin{figure}[t]
	\centering
	\hspace*{-2em}
	\begin{tikzpicture}[thick,scale=0.6, every node/.style={sloped,anchor=south,scale=0.8}]
	
	
	\node (l1) at (-2,0) [circ]{$x_3$};
	
	\node (l2) [circ, below left=1.4cm of l1]{$x_2$};
	\node (l3)[circ, below right=1.4cm of l1]{$x_1$};
	\draw[-] (l1) to node {} (l2);
	\draw[-] (l1) to node {} (l3);
	
	\node (l5)[circ, below left=1cm of l2]{$x_1$};
	\node (l4)[below right=1cm of l2]{$\ell_3$};
	\draw[-] (l2) to node {$\leq 1$} (l5);
	\draw[-] (l2) to node {$> 1$} (l4);
	
	\node (r4)[below left=1cm of l3]{$r_1$};
	\node (r5)[circ, below right=1cm of l3]{$x_2$};
	\node (r6)[below left=1cm of r5]{$r_2$};
	\node (r7)[below right=1cm of r5]{$r_3$};
	\draw[-] (r5) to node {$\leq 3$} (r6);
	\draw[-] (r5) to node {$> 3$} (r7);
	
	\node (l6)[below left=1cm of l3]{};
	\node (l7)[below right=1cm of l3]{};
	\draw[-] (l3) to node {$\leq 1$} (l6);
	\draw[-] (l3) to node {$> 1$} (l7);
	
	\node (l8)[below left=1cm of l5, scale=.9]{$\ell_1$};
	\node (l9)[below right=1cm of l5, scale=.9]{$\ell_2$};
	\draw[-] (l5) to node {$\leq 2$} (l8);
	\draw[-] (l5) to node {$> 2$} (l9);
	
	\tikzset{transition/.style = {rectangle, draw=black!50, fill=black!20, thick, minimum width=0.2cm,
			minimum height = 1cm},
	}
	\hskip0.25in
	\draw [fill=lightgray, opacity=0.2, thin] (-8,-10) rectangle node[color=black,opacity=1] {} 
	(-4,-7) ;
	\draw [fill=lightgray, opacity=0.2, thin] (-8,-11) rectangle node[color=black,opacity=1] {} 
	(-6,-10) ;
	\draw [fill=lightgray, opacity=0.2, thin] (-6,-11) rectangle node[color=black,opacity=1] {} 
	(-4,-10) ;
	
	\node[draw=none] at (-9.0,-10.9) {$x_2\leq 1$};
	\node[draw=none] at (-9.0,-9) {$x_2>1$};
	\node[draw=none] at (-7, -11.7) {$x_1\leq 2$};
	\node[draw=none] at (-5, -11.7) {$x_1> 2$};
	\node[draw=none] at (-7,-10.9) {$\ell_1$};
	\node[draw=none] at (-5,-10.9) {$\ell_2$};
	\node[draw=none] at (-6,-9) {$\ell_3$};
	
	\hskip0.7in
	\draw [fill=lightgray, opacity=0.2, thin] (-5.5,-11) rectangle node[color=black,opacity=1] {} 
	(-4.5,-7) ;
	\draw [fill=lightgray, opacity=0.2, thin] (-4.5,-11) rectangle node[color=black,opacity=1] {} 
	(-1.5,-8) ;
	\draw [fill=lightgray, opacity=0.2, thin] (-4.5,-8) rectangle node[color=black,opacity=1] {} 
	(-1.5,-7) ;
	
	\node[draw=none] at (-0.5,-10) {$x_2\leq 3$};
	\node[draw=none] at (-0.5,-7.8) {$x_2>3$};
	\node[draw=none] at (-5, -11.7) {$x_1\leq 1$};
	\node[draw=none] at (-3, -11.7) {$x_1> 1$};
	\node[draw=none] at (-5,-9.5) {$r_1$};
	\node[draw=none] at (-3,-10) {$r_2$};
	\node[draw=none] at (-3,-7.8) {$r_3$};

	\end{tikzpicture}
	\caption{Suppose we have three variables $(x_1,x_2,x_3)$ and when
		we split on reshaped variable $x_3$, the left and right subtrees and their corresponding cells 
		are 
		as shown above. By examination, any point that drops to $\ell_2$ can only travel to $r_2$ when 
		its 
		$x_3$ coordinate is increased. By this logic, the 
		exact estimator would use the six constraints $, \mu_{\ell_2}  \leq \mu_{r_2},\mu_{\ell_1}  \leq 
		\mu_{r_1}, \mu_{\ell_1}  \leq 
		\mu_{r_2},\mu_{\ell_3}  \leq \mu_{r_1},\mu_{\ell_3}  \leq 
		\mu_{r_2},\mu_{\ell_3}  \leq \mu_{r_3}$, whereas the over-constrained estimator would use all nine 
		pairwise constraints.} 
	\label{fig:ex}
	\vskip-1.5em
\end{figure}
\subsubsection{Exact Estimator}
\label{ex-sec}
Each leaf $\ell$ in a decision tree is a cell (or hyperrectangle) $C_\ell$ which is an intersection of 
intervals
\[
C_\ell = \bigcap_{j=1}^d \{ x : x_j \in I_j^\ell\}
\]

When we split on a shape-constrained variable $v$ with split-value $t$, each cell in the left subtree 
is of the form $C_l = \bar{C}_l \cap \{x :x_v \leq t\}$ and each cell in the right subtree is of the 
form $C_r = \bar{C}_r \cap \{x : x_v > t\}$.

For cells $l$ in the left subtree and $r$ in the right subtree, our goal is to constrain the 
corresponding leaf values
$\mu_l \leq \mu_r$ only when $\bar{C}_l \cap \bar{C}_r \neq \emptyset$. See Figure~(\ref{fig:ex}) for an 
illustration. We must devise an algorithm to find the intersecting cells $(l,r)$, and add each to a 
constraint set 
$E$. This can be done efficiently with an interval tree data structure. 

Assume there are $n$ unique leaves appearing in $E$. The exact estimator is obtained by solving the 
following optimization:
\begin{equation} \label{eq:exact}
\begin{aligned}
& \min_{(\hat{\mu}_\ell)_{\ell=1}^n}&&   \sum_{\ell=1}^{n} (\mu_\ell - \hat{\mu}_\ell)^2 \\  
& \text{subject to}  && \hat{\mu}_i \leq \hat{\mu}_j \; \forall \: 
(i,j) \in E\\
\end{aligned}
\end{equation}
where $\mu_\ell$ is the original value of leaf $\ell$. 

This is an instance of $L_2$ isotonic regression on a 
directed acyclic graph where each leaf value $\mu_\ell$ is a node, and each constraint in $E$ 
is an edge. With $n$ vertices and $m$ edges, the fastest known exact 
algorithm for this problem has time complexity $\Theta(n^4)$ \citep{spouge2003}, and the 
fastest known $\delta$-approximate algorithm has complexity $O(m^{1.5}\log^2 n \log \frac{n}{\delta})$ 
\citep{kyng2015}. 

With a corresponding change to the constraints in Equation~(\ref{eq:exact}), this approach extends 
naturally to convex regression trees. It can also be applied directly to probability trees for binary 
classification by reshaping the estimated probabilities in each leaf. 

\subsubsection{Over-constrained Estimator}
\label{oc-sec}
In this section, we propose an alternative estimator that can be more efficient to compute, 
depending on the 
tree structure. In our experiments below, we find that computing this estimator is always faster.  

Let $E_p$ denote the set of constraints that arise between leaf values under a shape-constrained split 
node $p$. By adding additional constraints to $E_p$, we can solve (\ref{eq:exact}) exactly 
for each shape-constrained split node in $O(n_p\log n_p)$ time, where $n_p$ is the number of leaves 
under 
$p$. 

In this setting, each shape-constrained split node gives rise to an 
independent optimization involving its leaves. Due to transitivity, we can solve these optimizations 
sequentially in reverse (bottom-up) level-order on the tree.  

Let $n_p$ denote the number of leaves under node $p$. For each node $p$ that is split on a 
shape-constrained variable, the over-constrained estimator solves 
the following max-min problem:
\begin{equation} \label{eq:oc}
\begin{aligned}
& \min_{(\hat{\mu}_\ell)_{\ell=1}^{n_p}}&&   \sum_{\ell=1}^{n_p} (\mu_\ell - \hat{\mu}_\ell)^2 \\  
& \text{subject to}  && \max_{\ell \in \text{left}(p)} 
\hat{\mu}_\ell \leq \min_{r \in \text{right}(p)} \hat{\mu}_r \\
\end{aligned}
\end{equation}
where left($p$) denotes all leaves in the left subtree of $p$ and right($p$) denotes all leaves in the 
right subtree. 

This is equivalent to adding an edge $(\ell, r)$ to $E$ for every pair of leaves such that $\ell$ is in 
left($p$) and $r$ is in right($p$). All such pairs do not necessarily exist in $E$ for the 
exact estimator; see Figure~(\ref{fig:ex}). For each shape-constrained split, (\ref{eq:oc}) is an instance 
of $L_2$ isotonic 
regression on a complete directed bipartite graph. 

For a given shape-constrained split node $p$, let $\ell = (\ell_1, \ell_2, \dots, \ell_{n_1})$ be the values of 
the 
leaves in its left subtree, and $r = (r_1, r_2, \dots, r_{n_2})$ be the values of the leaves in 
its right subtree, indexed so that $\ell_1 \leq \dots \leq \ell_{n_1}$ 
and $r_1 \leq \dots \leq r_{n_2}$. Then the max-min problem (\ref{eq:oc}) is equivalent to:
\begin{equation} 
\begin{aligned}
&\min_{\tilde{\ell}, \tilde{r}}&&  \sum_{i=1}^{n_1} (\ell_i - 
\tilde{\ell}_i)^2 + \sum_{i=1}^{n_2} (r_i - \tilde{r}_i)^2   \\  
& \text{subject to}  && \tilde{\ell}_1 \leq \tilde{\ell}_2 \leq ... \leq \tilde{\ell}_{n_1} \leq \tilde{r}_1 \leq \dots 
\leq \tilde{r}_{n_2}\\
\end{aligned}
\end{equation}
\begin{figure*}[b!]
	\centering
	\begin{subfigure}{0.49\textwidth}
		\centering
		\includegraphics[scale=0.4]{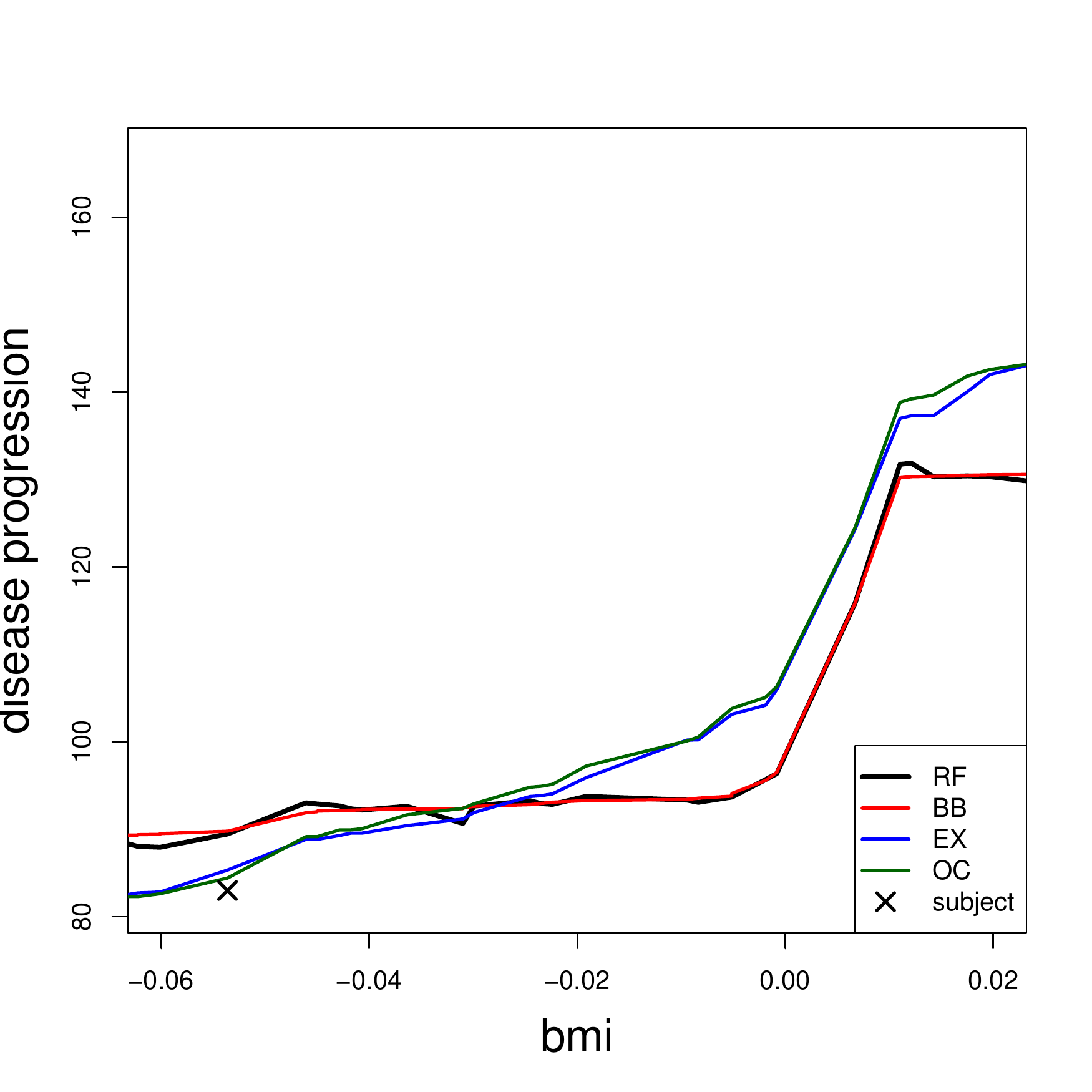}
		\caption{Diabetes dataset.}
		\label{fig:diabetes}
	\end{subfigure}
	\begin{subfigure}{0.49\textwidth}
		\centering
		\includegraphics[scale=0.4]{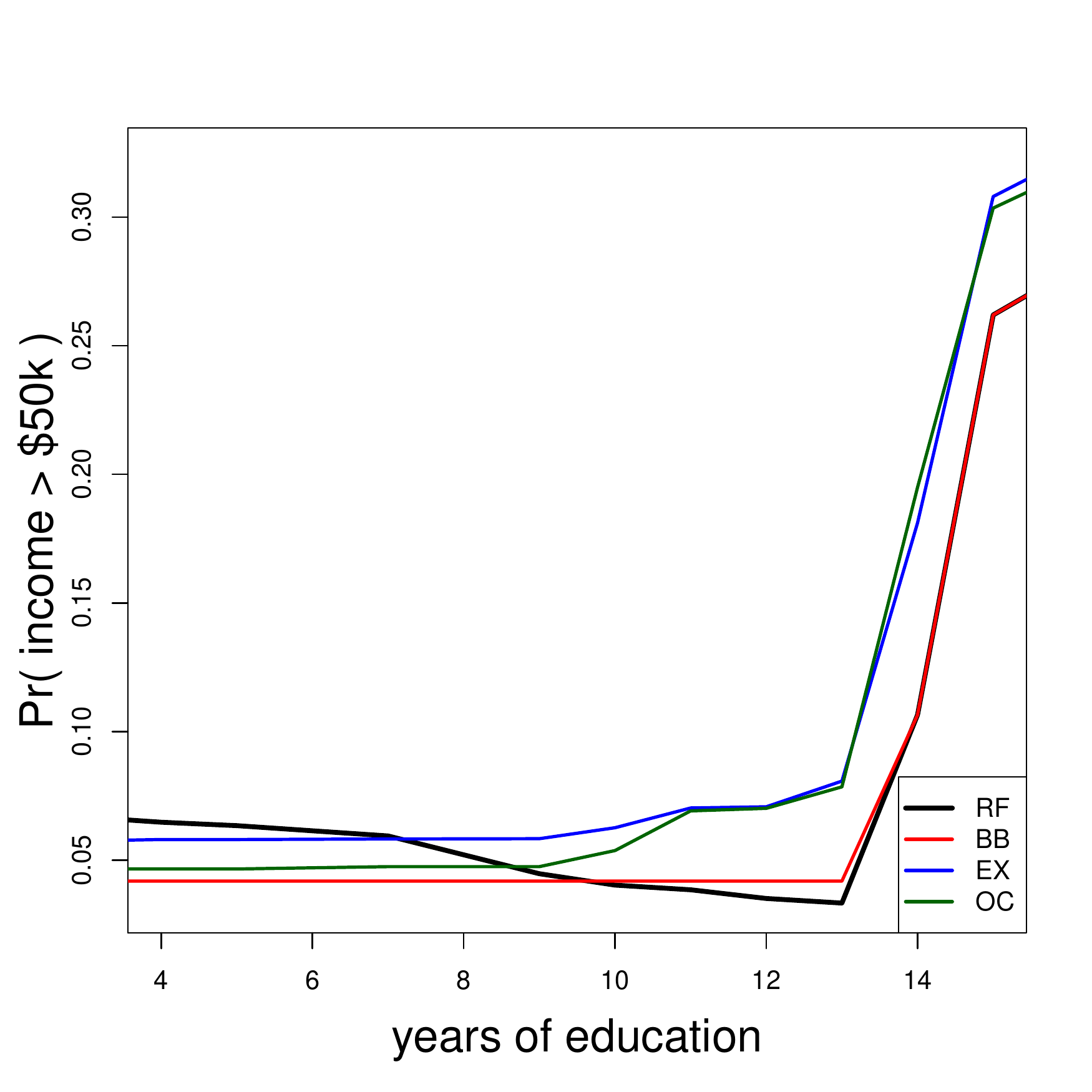}
		\caption{Adult dataset.}
		\label{fig:adult}
	\end{subfigure}
	\caption{Illustrating the result of reshaping. We choose a test point at random and make predictions 
		with each model as we vary the predictor variable on the 
		$x$-axis, holding all other 
		variables constant. We see in both cases that the original RF predictions are not 
		monotone-increasing. The 
		Diabetes 
		plot (\ref{fig:diabetes}) also shows the true value of the chosen data point with an X. }
	\label{fig:preds}
\end{figure*}

The solution to this 
optimization is of the form $\tilde{\ell}_i = \min (c, \ell_i)$ and $\tilde{r}_i  = \max (c, r_i)$, for some 
constant $c$. Given the two sorted vectors $\ell$ and $r$, the optimization becomes:
\begin{equation*}
\min_c \sum_{i=1}^{n_1} (\ell_i - \min(c, \ell_i))^2 + \sum_{i=1}^{n_2} (r_i - \max(c, r_i))^2  
\end{equation*}

This objective is convex and differentiable in $c$. Similar to the black box reshaping method, we can 
compute the derivatives at the values of the 
data and find where it flips sign, then compute the minimizer in the 
corresponding segment. This takes $O(n)$ time where $n = n_1 + n_2$, the number of leaves under 
the shape-constrained split. With sorting, the over-constrained estimator can be computed in $O(n\log 
n)$ time for each shape-constrained split node. 

We apply this procedure sequentially on the leaves of every shape-constrained node in reverse 
level-order on the tree. 

\section{Experiments} \label{sec:exp}
\begin{table}[h]
	\caption{Acronyms used for reshaping methods. }
	\label{acronym-table}
	\vskip 0.15in
	\begin{center}
		\begin{small}
			\begin{sc}
				\begin{tabular}{lcr}
					\toprule
					Method & Acronym  & Sec\\
					\midrule
					Random Forest & RF & - \\
					Black Box Reshaping & BB &  \ref{sec:bb} \\
					Exact Estimator & EX  & \ref{ex-sec}  \\
					Over-constrained Estimator & OC & \ref{oc-sec}\\
					\bottomrule
				\end{tabular}
			\end{sc}
		\end{small}
	\end{center}
	\vskip -0.1in
\end{table}

We apply the reshaping methods described above to two regression tasks and two binary 
classification 
tasks. We show that reshaping allows us to enforce shape constraints 
without compromising predictive accuracy. For convenience, we use the acronyms in 
Table~(\ref{acronym-table}) to refer to each method. 

The BB method was implemented in R, and the OC and 
EX 
methods were implemented in R and 
C++, extending the R package \texttt{ranger}  
\citep{wright2017}. The exact estimator from Section~(\ref{ex-sec}) is computed using the MOSEK C++ 
package \citep{mosek}. 

For binary classification, we use the probability tree implementation found in 
\texttt{ranger}, enforcing monotonicity of the probability of a positive classification with respect to the 
chosen predictors. For the purposes of these experiments, black box reshaping is applied to a 
traditional random 
forest. The random forest was fit with the default settings found in \texttt{ranger}. 

We apply 5-fold cross validation on all four tasks and present the results under the relevant 
performance metrics in Table~(\ref{table:results}).

\subsection{Diabetes Dataset}
\setlength{\tabcolsep}{12pt}
\begin{table*}[t!]
	\centering
	\caption{Experimental results of 5-fold cross-validation. Accuracy is measured by mean squared 
		error (Diabetes), mean absolute percent error (Zillow), and 
		classification accuracy (Adult, Spam). Standard errors are shows in parentheses.}
	\label{table:results}
	\vskip 0.15in
	\begin{center}
		\begin{small}
			\begin{sc}
				\begin{tabular}{lcccc}
					\toprule
					Method& Diabetes & Zillow & Adult & Spam  \\
					\midrule
					RF    &  3209 (377) & 2.594\% (0.1\%)  &87.0\% (1.7\%)& 95.4\% (1.3\%) \\
					BB    &  3210 (390) & 2.594\% (0.1\%) & 87.1\% (1.8\%)& 95.4\% (1.3\%)\\
					EX    &  3154 (353) &  -                     &  86.8\% (1.5\%) & 95.3\% (1.2\%) \\
					OC    &  3155 (350) &2.588\% (0.1\%)&   87.0\% (1.5\%)&   95.3\% (1.2\%)  \\
					\bottomrule  
				\end{tabular}
			\end{sc}
		\end{small}
	\end{center}
	\vskip -0.1in
\end{table*}
\setlength{\tabcolsep}{6pt}
The diabetes dataset \citep{efron2004} consists of ten physiological baseline variables, age, 
sex, body mass index, average blood pressure, and six blood serum measurements, for each of 442 
patients. The response is a quantitative measure of disease progression measured one year after 
baseline. 

Holding all other variables constant, we might expect disease progression to be monotonically 
increasing in body mass index \citep{ganz2014}. We estimate a random forest and apply our 
reshaping techniques, then make predictions for a random test subject as we vary the body mass index 
predictor variable. The results shown in Figure~(\ref{fig:diabetes}) illustrate the effect of reshaping on 
the predictions. 

We use mean squared error to measure accuracy. The results in Table~({\ref{table:results}})
indicate that the prediction accuracy of all four estimators is approximately the same. 

\subsection{Zillow Dataset}
In this section, the regression task is to predict real estate sales prices using property information. The 
data were obtained from Zillow, an online real estate database company. For each of 
206,820 properties, we are given the list price, number of bedrooms and bathrooms, square 
footage,  build decade, sale year, major renovation year (if any), city, and metropolitan area. The 
response is the actual sale price of the home. 

We reshape to enforce monotonicity of the sale price with respect to the list price. Due to the size of 
the constraint set, this problem becomes intractable for MOSEK; the results for the EX method are 
omitted. An interesting  direction for future work is to investigate 
more efficient algorithms for this method.  

Following reported results from Zillow, we use mean absolute percent error (MAPE) as our measure of 
accuracy. For an estimate $\hat{y}$ of the true value $y$, the APE is $|\hat{y} - y|/y$.

The results in Table~(\ref{table:results}) show that the performance across all estimators is 
indistinguishable. 

\subsection{Adult Dataset}
We apply the reshaping techniques to the binary classification task found in the Adult dataset 
\cite{lichman2013}. The task is to predict whether an individual's income is less than or greater than 
\$50,000. Following the experiments performed in \citet{fard2016} and \citet{you2017}, we apply 
monotonic reshaping to four variables: capital gain, weekly hours of work, education level, and the 
gender wage gap. 

We illustrate the effect of reshaping on the predictions in Figure~(\ref{fig:adult}). The results in 
Table~(\ref{table:results}) show that we achieve similar test set 
accuracy before and after reshaping the random forest.

\subsection{Spambase Dataset}
Finally, we apply reshaping to classify whether an email is spam or not. The Spambase dataset 
\citep{lichman2013} 
contains 4,601 emails each with 57 predictors. There are 48 word frequency predictors, 6 character 
frequency predictors, and 3 predictors related to the number of capital letters appearing in the email. 

That data were collected by Hewlett-Packard labs and donated by George Forman. One of the 
predictors is the frequency of the word ``george", typically assumed to be an indicator of non-spam 
for this dataset. 
We reshape the predictions to enforce the 
probability of being classified as spam to be monotonically decreasing in the frequency of words 
``george" 
and ``hp". 

The results in Table~(\ref{table:results}) again show similar performance across all methods. 

\section{Discussion}
We presented two strategies for prediction rule reshaping. We developed 
efficient algorithms to compute the reshaped estimators, and illustrated their properties on four 
datasets. Both approaches can be viewed as frameworks for developing more sophisticated 
reshaping techniques. 

There are several ways that this work can be extended. Extensions to 
the black box method include adaptively combining rearrangements and isotonization
\citep{chernozhukov2009}, and considering a weighted objective function to account for the distance 
between test points. 

In general, the random forest reshaping method can be viewed as operating on pre-trained parameters 
of a specific model. Applying this line of thinking to gradient boosted trees, 
deep learning methods, and other machine learning techniques could yield useful variants of this 
approach. 

And finally, while practitioners might require certain shape-constraints on their predictions, many 
scientific applications also require inferential quantities, such as
confidence intervals and confidence bands. Developing inferential procedures for 
reshaped predictions, similar to \citet{chernozhukov2010} for rearrangements and  \citet{athey2016} 
for 
random forests, would yield interpretable predictions along with useful measures of uncertainty. 
\section{Acknowledgments}
Research supported in part by ONR grant N00014-12-1-0762, NSF grants DMS-1513594 and 
DMS-1654076, and an Alfred P. Sloan fellowship.

\setlength{\bibsep}{6pt}
\bibliography{reshape}
\bibliographystyle{icml2018}
\end{document}